\documentclass[journal,twoside]{IEEEtran}
\usepackage[noadjust]{cite}

\usepackage{color,soul}
\soulregister\cite7 
\soulregister\citep7 
\soulregister\citet7 
\soulregister\ref7 
\soulregister\pageref7 

\ifCLASSINFOpdf
  \usepackage[pdftex]{graphicx}
  \graphicspath{{./pdf/}{./jpeg/}}
  \DeclareGraphicsExtensions{.pdf,.jpeg,.PNG}
\else
  \usepackage[dvips]{graphicx}
  \graphicspath{{./eps/}}
  \DeclareGraphicsExtensions{.eps}
\fi
\usepackage{xcolor}
\usepackage[switch]{lineno}
\usepackage{amsmath,amssymb,amsthm,mathtools}
\usepackage{algorithm}
\usepackage{algorithmic}
\usepackage{bm}
\usepackage{multirow}

\usepackage{makecell}

\begin{document}
%
\title{
HandCept: A Visual-Inertial Fusion Framework for Accurate Proprioception in Dexterous Hands
} 
%
%
\author{Huang Junda$^{1}$, Honghao Guo$^{1}$, Hao Wu$^{2}$, Zhengyang Liu$^{3}$, Marcelo H Ang Jr$^{2}$, Jianshu Zhou$^{2}$

        
\thanks{ This work is supported by the CUHK T Stone Robotics Institute, in part by the InnoHK initiative of the Innovation and Technology Commission of the Hong Kong Special Administrative Region Government via the
Hong Kong Centre for Logistics Robotics, and in part by the HK RGC under GRF-14207423.} 
\thanks{$^{1}$Authors with the Department of Mechanical and Automation Engineering, The Chinese University of Hong Kong.}
\thanks{$^{2}$Authors with the Department of Mechanical Engineering, National University of Singapore.}
\thanks{$^{3}$Independent researcher.}
}

\maketitle


\begin{abstract}

As robotics progresses toward general manipulation, dexterous hands are becoming increasingly critical. However, proprioception in dexterous hands remains a bottleneck due to limitations in volume and generality. In this work, we present HandCept, the first visual-inertial proprioception framework designed to overcome the challenges of traditional joint angle estimation methods for dexterous hands. HandCept addresses the difficulty of achieving accurate and robust joint angle estimation in dynamic environments where both visual and inertial measurements are prone to noise and drift. It leverages a zero-shot learning approach using a wrist-mounted RGB-D camera and 9-axis IMUs, fused in real time via a latency-free Extended Kalman Filter (EKF). Our results show that HandCept achieves joint angle estimation errors generally between $2^{\circ}$ and $4^{\circ}$ without observable drift, outperforming visual-only and inertial-only methods. Furthermore, we validate the stability and uniformity of the IMU system, demonstrating that a common base frame across IMUs simplifies system calibration. To support sim-to-real transfer, we also open-source our high-fidelity rendering pipeline, which is essential for training without real-world ground truth. This work offers a robust, generalizable solution for proprioception in dexterous hands, with significant implications for robotic manipulation and human-robot interaction. https://github.com/huangjund/blenderYCB

\end{abstract}


%

\section{Introduction}

\IEEEPARstart{R}{}obots are increasingly being deployed in a wide range of diverse tasks and are expected to operate in unstructured environments such as homes, shopping malls, and hospitals \cite{zeng2023large,laschi2016soft}. Performing these tasks often involves manipulation—either bimanual or in-hand dexterous manipulation—to interact with the environment. The end-effector plays a fundamental role in enabling such robotic manipulation \cite{mason2018toward,billard2019trends,cui2021toward}. 

Recent advances in dexterous hands \cite{dollar2014special} and two-finger grippers \cite{chi2024universal} across various applications have demonstrated the potential and importance of further developing dexterous hands. A wide variety of dexterous hand designs have emerged, incorporating diverse structural topologies such as roller joints \cite{yuan2020design}, prismatic joints\cite{zhou2024dexterous}, compliant joints \cite{odhner2015stable}, soft joints \cite{shepherd2011multigait}, functional joints \cite{stuart2015suction}, and transformable joints \cite{zhou2025prismatic}. Notably, revolute joint chains can integrate compliant, universal, or transformable joints, as their joint angles can still be inferred from the 3D poses of associated rigid  \cite{huang2025dih,zhou2025unified}. However, due to limitations in proprioceptive frameworks, many innovative dexterous hand designs remain confined to demonstrations of structural dexterity or teleoperated in open-loop control \cite{zhou2019soft, yuan2020design}.

\begin{figure}[!t] 
    \centering
    \includegraphics[width=\linewidth]{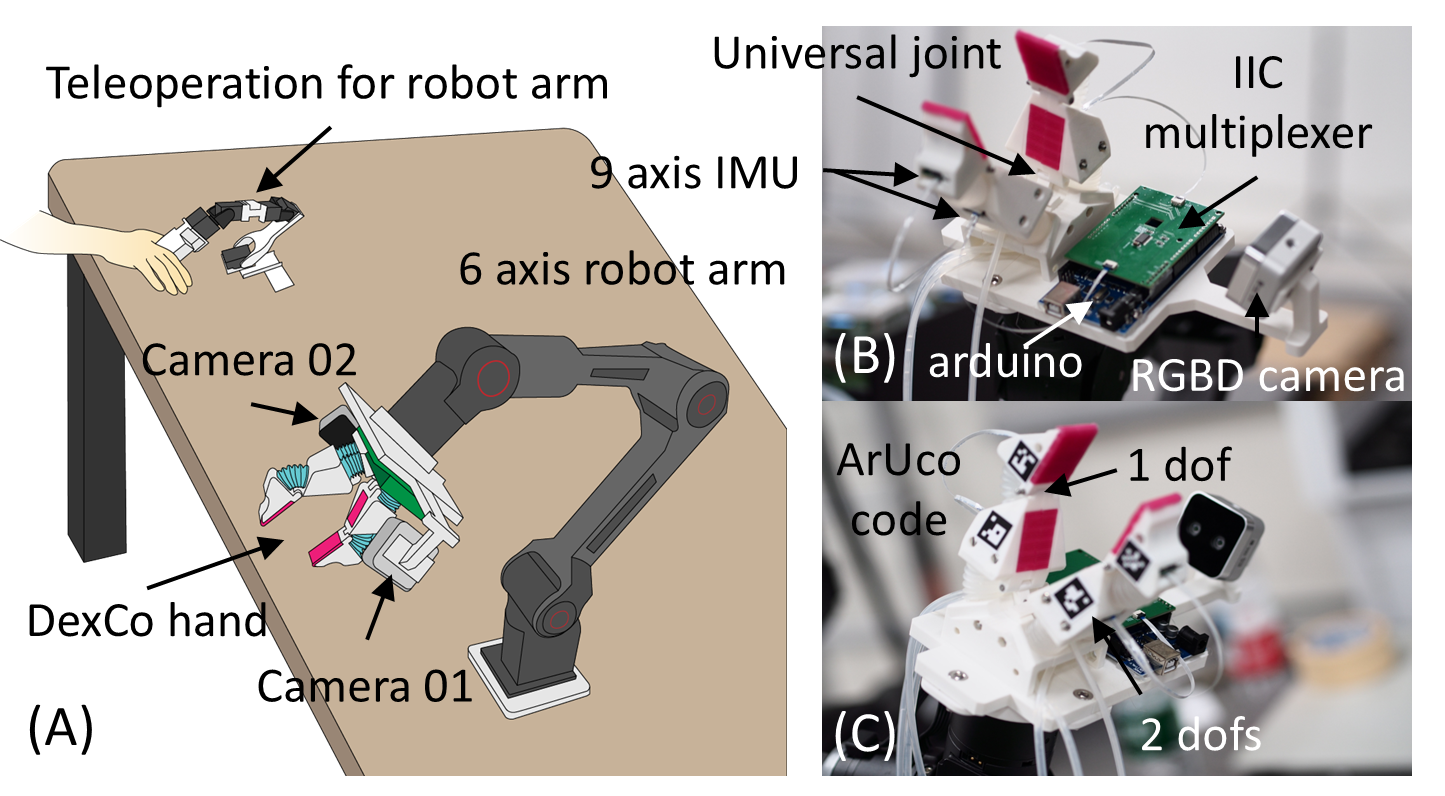}
    \caption{
        (A) Overview of the HandCept implementation on the DexCo hand \cite{zhou2024dexterous}. (B) and (C) show the physical DexCo hand (with universal joint); ArUco markers are used in the experiments to obtain ground-truth poses for each link.
    }
    \label{fig:overview}
\end{figure}

Traditional proprioception approaches in dexterous hands generally fall into two categories: direct and indirect methods. Direct methods (e.g. encoder-based) \cite{shaw2023leap,chen2023visual, yuan2020design} offer accurate and direct joint angle feedback, but their physical size limits integration into compact hand architectures and restricts the number of degrees of freedom per joint. Indirect methods (e.g. tendon-based) \cite{dollar2014special,odhner2014compliant,quigley2014mechatronic,ciocarlie2014velo}, while more flexible in mechanical design, often suffer from poor stability, calibration complexity, and increased maintenance. 

Several novel sensing techniques have also been explored, such as optical fibers \cite{riza2020fbg, kang2023soft}, inductive sensors \cite{huang2022modular}, capacitive sensors \cite{dawood2020silicone}, magnetic field sensors \cite{mitchell2021fast}, vision-based estimation \cite{cong2022reinforcement}, and fluidic sensors \cite{helps2018proprioceptive, alatorre2021continuum}. However, these alternatives still face limitations in terms of robustness, accuracy, and ease of integration in practical implementation.

In parallel, visual-inertial perception has been widely adopted in fields such as autonomous locomotion and virtual reality. For mobile robots, visual-inertial odometry techniques focus on full-body localization \cite{teixeira2018vi, chen2023stereo}. In VR and AR applications, hand tracking systems \cite{guedri2021finger,li2023visual} use combined sensory input to estimate hand poses, typically prioritizing generalization and user interaction. Nevertheless, these systems often lack the precision, reliability, and long-term robustness required in closed-loop robotic manipulation and thus are not directly transferable to dexterous hands operating under dynamic and task-critical conditions.

To address proprioception challenges in dexterous hands, we present \textbf{HandCept}, a visual-inertial proprioception framework for joint angle estimation in rigid-link dexterous hands. HandCept combines a miniaturized 9-axis IMU module (12 mm × 15 mm) with scalable serial and parallel integration, and a kinematically constrained estimation algorithm that fuses visual and inertial pose observations. Visual poses are obtained from a wrist-mounted RGB-D camera and a zero-shot pose estimation network trained entirely on high-fidelity synthetic data generated with a Blender-based pipeline, eliminating the need for real-world data collection while enabling sim-to-real transfer. A latency-aware Extended Kalman Filter further integrates asynchronous visual and inertial streams for robust real-time pose estimation, which is then converted into joint angles. Experiments demonstrate the feasibility, robustness, and long-term precision of HandCept, showing superior performance over single-modality baselines and an average joint angle error of $2^{\circ}$ -- $4^{\circ}$. We further validate the stability and uniformity of the IMU system and show that a shared reference frame across multiple IMUs simplifies calibration.

The remainder of this paper is organized as follows: Section II provides an overview of the HandCept framework. Section III details the rigid-body pose estimation methods. Section IV introduces the EKF-based sensor fusion algorithm. Section V presents the experimental evaluation. Section VI concludes the paper and outlines future directions.

\begin{figure*}[!t] 
    \centering
    \includegraphics[width=\linewidth]{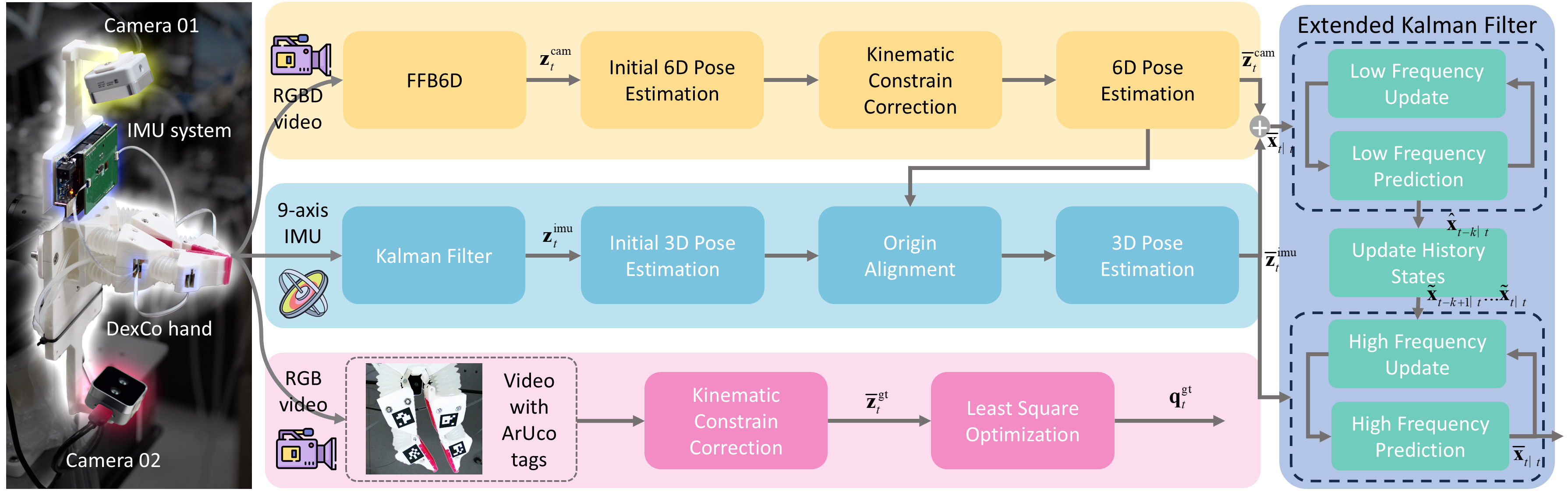}
    \caption{
       Algorithmic pipeline of HandCept. The ArUco code-based ground truth system is used solely for experimental validation. HandCept estimates the joint angles of the DexCo hand by fusing visual and inertial data streams. The pose estimates are refined with kinematic constraints before being fused using an Extended Kalman Filter (EKF). Due to the latency in visual processing, each visual-inertial update is used to correct the estimate at time $t+1$, starting from $k$ previous steps (i.e., at $t-k$). Meanwhile, IMU data provides real-time updates via the EKF.
    }
    \label{fig:framework}
\end{figure*}

\section{HandCept Framework Overview}

HandCept is a proprioception framework designed to estimate joint angles of dexterous robotic hands. HandCept adopts a model-based approach to proprioception, as illustrated in Fig.~\ref{fig:framework}. The system leverages two sensory modalities—inertial and visual—that together contribute to estimating the pose of each rigid link within the dexterous hand. The inertial component, derived from multiple 9-axis IMUs, captures the 3D orientation (rotation) of each link. The visual component, provided by a wrist-mounted RGB-D camera, estimates the full 6D pose (rotation and translation) of the links. The design and implementation details of the 9-axis IMUs are elaborated in a subsequent section.

The visual estimation component is based on FFB6D, an instance-level object pose estimation algorithm that produces segmentation masks, keypoints, and 6D poses for all objects within an input image. Given the lack of accessible ground-truth data for many dexterous hands, we developed a dedicated Blender-based simulation pipeline to generate high-quality synthetic training data, enabling zero-shot sim-to-real transfer for the FFB6D model. The initial 6D pose estimates from FFB6D are further refined using kinematic constraints derived from the known topology of the hand, thereby improving the accuracy of the visual estimates. These refined visual estimates are then used to calibrate the inertial measurements, aligning both modalities to a unified reference frame. Finally, the visual and inertial data are fused using an Extended Kalman Filter (EKF), resulting in robust, drift-free, and accurate 3D pose estimates for each link. These fused 3D poses are used to compute the joint angles of the dexterous hand, thus enabling accurate proprioception.

\section{Rigid Body Pose Estimation}
\label{sec:pose_estimation}

In this section, we outline the methods employed to estimate the poses from both inertial and visual systems. Unlike general-purpose algorithms designed to estimate the 6D pose of arbitrary objects, robotic systems typically consist of tree-like structures, where kinematic constraints among connected rigid bodies play a critical role. Consequently, incorporating these kinematic constraints into the pose estimation process is essential. It is important to note that this work focuses exclusively on systems with revolute joint chains; other joint types (e.g., prismatic or soft joints) are not considered.

\subsection{Inertial-based SO3 Estimation}
\begin{figure}[!t] 
    \centering
    \includegraphics[width=\linewidth]{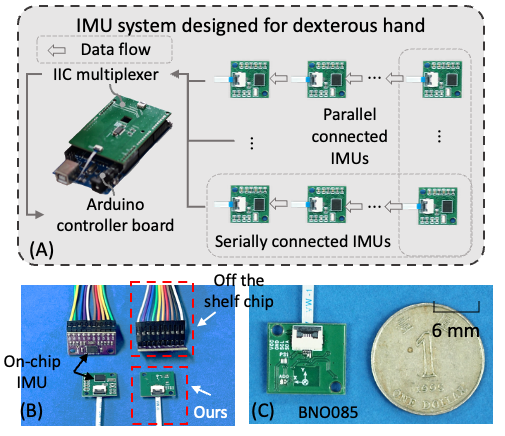}
    \caption{
        (A) Architecture of the modular IMU system. Using the I²C protocol, multiple 9-axis IMUs can be connected in series to expand the system. The I²C multiplexer supports parallel connections, enabling applications such as five-fingered hands. An Arduino collects and transmits the data to a computer via a serial interface. (B) and (C) are front and back views of the designed IMU module—the smallest known IMU boards-compared with the smallest off-the-shelf BNO085 IMU board, offering low cost and high accuracy for general usage. 
    }
    \label{fig:imu}
\end{figure}

The IMU system is designed to support both serial and parallel extensibility. Using the I²C protocol, IMU PCBs can be serially connected to extend the system, while a custom-designed I²C multiplexer enables parallel connections to accommodate different numbers of dexterous fingers or limbs (Fig.\ref{fig:imu}A). This IMU system is also the smallest known design, facilitating easy integration into dexterous hands while maintaining a real-time update frequency of $\geq$200Hz (Fig.~\ref{fig:imu}B). Utilizing 9-axis IMUs is essential for pose estimation, as it reduces the calibration burden. Given the compactness of dexterous hands, the magnetic field can be assumed uniform across all IMUs, allowing the use of a common base frame across different modules—a condition verified in our experiments.

For the pose estimation algorithm, we leverage the inherent stability of the Kalman filter, which are widely adopted for 9-axis IMU-based orientation estimation. As a result, the direct use of raw IMU data (acceleration, magnet, and angular velocity) is typically unnecessary for most orientation applications. In our framework, a Kalman filter is employed to estimate the 3D orientation (SO(3)) of each link reliably, thereby providing robust inertial-based pose measurements, denoted as $\mathbf{z}_t^{\mathrm{imu}}$.

Let $\{IMU_i\}$, $\{IMU_b\}$, $\{b\}$, $\{ee\}$, and $\{L_i\}$ denote the $i$-th IMU frame, the common IMU base frame, the robot arm's base frame, the end-effector frame, and the $i$-th finger link's frame, respectively. Let $^aT_b$ represent the homogeneous transformation matrix from frame $a$ to frame $b$. As the base frame of the IMU is fixed relative to the arm's base frame, the estimated transformation from the $i$-th IMU frame to the end-effector frame is expressed as:
\begin{equation}
    ^{IMU_i}\hat{T}_{ee} = ^{IMU_i}\hat{T}_{IMU_b}(\mathbf{z}^{\mathrm{imu}}_t) {}^{IMU_b}T_b{}^bT_{ee}(\boldsymbol{\theta})
\end{equation}
where $\hat{\left(\cdot\right)}$ means the measured value, $\boldsymbol{\theta}$ is the joint angles of the robot arm. The right hand side transformations are known.

\subsection{Visual-based SE3 Estimation}
\begin{figure*}[!t] 
    \centering
    \includegraphics[width=\linewidth]{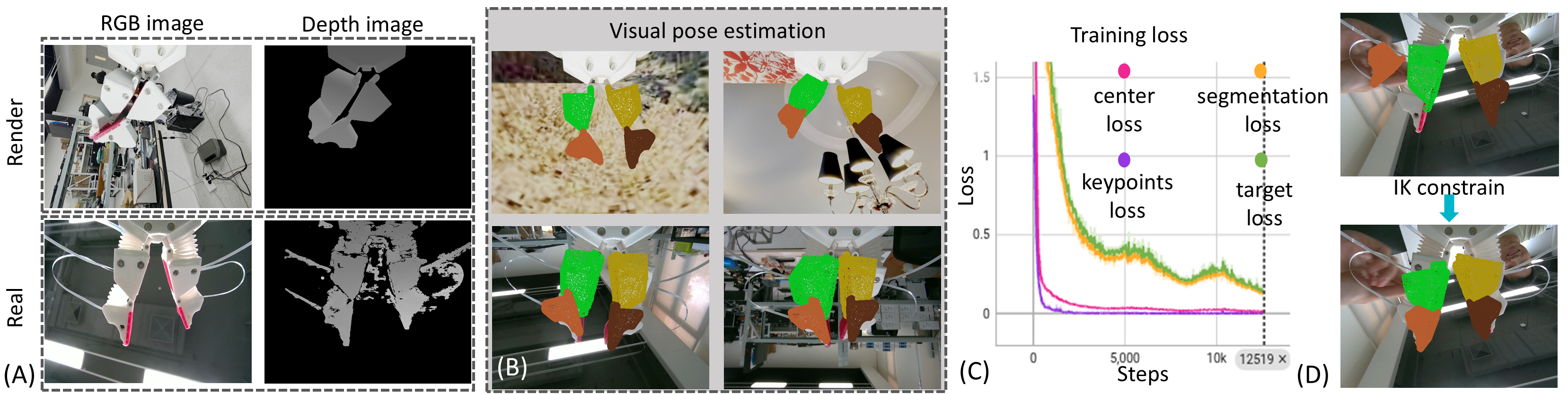}
    \caption{
        (A) Comparison between high-fidelity rendered RGB-D images and real-world images. (B) Zero-shot 6D pose estimation results: the first row shows predictions on rendered images, and the second row shows predictions on real images across varying end-effector poses and hand configurations. (C) Training loss progression across four loss components. (D) Improved 6D pose estimation by enforcing kinematic constraints. Human-hand interference introduces inaccurate pose estimates, which are corrected by removing undesired rotations and translations through kinematic constraints.
    }
    \label{fig:render}
\end{figure*}

The visual system leverages a wrist-mounted RGB-D camera to provide full 6D pose (SE(3)) estimation for each finger link. Considering it is hard for the HandCept-targeted dexterous hand to get the ground truth SE3 of each link in real scenario, we decided to purely use synthetic data for zero-shot training. The rendering algorithm based on blender performs high fidelity rendering by using both the UV mapping and real panoramic background (Fig. \ref{fig:render}A), which is also inspired by the fast growing real2sim tech. However, it's obvious that the real-world images are clutter and noisy, while we didn't add noise for sim2real transfer in this study.

The FFB6D \cite{he2021ffb6d} is used as the segmentation and 6D pose estimation algorithm, which provides instance-level prediction. A sum loss, including the segmentation loss (Focal Loss), center point loss (L1 Loss), and keypoints loss (L1 Loss), are shown in the Fig. \ref{fig:render}C in the training process. As shown in the Fig. \ref{fig:render}B, the quanlified results of visual estimations on the rendered images are highly accurate under different backgrounds, while the real images have some error.

\subsection{Kinematic Constraints}
In robotic systems, the interdependencies among the rigid bodies are governed by specific kinematic constraints determined by the revolute joint chain topology. Suppose $^aR_b \in SO(3)$ is the rotation matrix from frame a to frame b. Appilying least square method
\begin{equation}
\min_{R_i} \frac{1}{2}\left\| \tilde{R} - \overline{R} \cdot R_i \right\|_F^2,
\end{equation}
where $i \in \{x, y, z\}$ and $R_i$ represents the rotation matrix around axis $i$, $\tilde{R}$ is the observed rotation, $\overline{R}$ is the rotation matrix considering kinematic constraints, $\|\cdot\|_F$ denotes the Frobenius norm. The extra rotation $R_i$ can be therefore be removed from the measured $\mathbf{z}_t^{\mathrm{imu}}$ (Fig. \ref{fig:framework}).

These constraints are incorporated into the pose estimation process to enforce consistency across the chain, thereby improving the overall accuracy and stability of the estimated 6D poses (Fig. \ref{fig:render}D).

\section{Fusion using Latency-Free EKF}
\label{sec:fusion}

The Extended Kalman Filter is a powerful tool for state estimation in non-linear systems. It operates through a two-step process: prediction and update. In the prediction step, the EKF uses a system model to project the current state estimate and its associated covariance forward in time. This prediction is based on the system's dynamics and is subject to process noise, which accounts for uncertainties in the model and the process itself. Following the prediction, the update step incorporates new measurements to correct the predicted state and covariance. This correction is achieved by calculating the Kalman gain, which determines the weight given to the new measurement based on the estimated uncertainties in both the prediction and the measurement (represented by the measurement noise). The difference between the actual measurement and the predicted measurement based on the current state estimate is then used to refine the state and reduce the uncertainty in the estimate. A key aspect of the EKF is its ability to handle non-linear systems by linearizing the non-linear process and measurement models around the current state estimate using Jacobian matrices. This linearization allows the application of Kalman filter principles to non-linear problems. We suppose $\hat{\left(\cdot\right)}$ as the visual-inertial fusion related values, $\tilde{\left(\cdot\right)}$ as the only inertial related values.

\subsection{Extended Kalman Filter}

To effectively estimate 3D rotation using an EKF, a suitable representation of the object's orientation must be chosen. Several options are available, including Euler angle, quaternion, and rotation matrix. Quaternions, specifically unit quaternions, offer a more compact and robust representation for 3D rotations, as well as avoiding gimbal lock and allowing for smooth interpolation.

Let the state vector at time $t$ be:
\begin{equation}
    \mathbf{x}_t=\left[\begin{array}{l}
\mathbf{q}_t \\
\mathbf{b}_t
\end{array}\right]
\end{equation}
where $\mathbf{q}_t=\left[q_w, q_x, q_y, q_z\right]^{\top}$ is the unit quaternion representing the link's orientation, $\mathbf{b}_t$ is the bias, which follows a random process.

Since the IMU gives orientation estimates, we assume the system follows a random walk for orientation
\begin{equation}
\begin{gathered}
    \mathbf{q}_{t+1}=\mathbf{q}_t \otimes \delta \boldsymbol{q}_t\\
    \mathbf{b}_{t+1}=\mathbf{b}_t+\boldsymbol{\eta}_t^b
\end{gathered}
\end{equation}
where $\boldsymbol{\delta} \boldsymbol{q}_t \approx\left[1, \frac{1}{2} \boldsymbol{\omega}_t \Delta t\right]$ is a small rotation quaternion representing random motion (as noise), $\boldsymbol{\eta}_t^b \sim \mathcal{N}\left(\mathbf{0}, Q_b\right)$ is the bias noise, $\otimes$ is the quaternion product. For simplicity, since we don't use the IMU's angular velocity, the model can be represented as
\begin{equation}
    \mathbf{q}_{t+1}=\mathbf{q}_t \otimes \mathbf{q}_{\text {noise }}
\end{equation}
which can be linearized to
\begin{equation}
\mathbf{x}_{t+1}=f\left(\mathbf{x}_t\right)+\mathbf{w}_t=\left[\begin{array}{c}
\mathbf{q}_t \otimes \delta \mathbf{q}_t \\
\mathbf{b}_t
\end{array}\right]+\mathbf{w}_t
\end{equation}
with $\mathbf{w}_t = \left[q_{\mathrm{noise},\boldsymbol{\eta}_t^b }\right]$, where $f\left(\mathbf{x}_t\right)=\mathbf{x}_{t}$, ${\mathbf{w}}_t$ is the drifting noise. The process noise covariance is
\begin{equation}
    Q_t=\left[\begin{array}{cc}
Q_q & 0 \\
0 & Q_b
\end{array}\right]
\end{equation}
where $Q_q$ and $Q_b$ are the quaternion propagation noise and bias noise respectively. It needs to be tuned for a best fitting.

\textbf{Measurement Models} There are two types of observations, which is IMU measurement model
\begin{equation}
    \begin{gathered}
    \mathbf{z}_t^{\mathrm{imu}}=h^{\mathrm{imu}}\left(\mathbf{x}_t\right)+\mathbf{v}_t^{\mathrm{imu}}
    \end{gathered}
\end{equation}
with
\begin{equation}
    h^{\mathrm{imu}}\left(\mathbf{x}_t\right)=\mathbf{q}_t \otimes \delta \mathbf{q}\left(\mathbf{b}_t\right)
\end{equation}
where $\delta \mathbf{q}\left(\mathbf{b}_t\right)$ represents a small rotation corresponding to the bias $\mathbf{b}_t$. For small angles, we approximate as
\begin{equation}
    \delta \mathbf{q}\left(\mathbf{b}_t\right) \approx\left[1, \frac{1}{2} \mathbf{b}_t\right]
\end{equation}

The camera measurement model
\begin{equation}
\mathbf{z}_t^{\mathrm{cam}}=h^{\mathrm{cam}}\left(\mathbf{x}_t\right)+\mathbf{v}_t^{\mathrm{cam}}=\mathbf{q}_t+\mathbf{v}_t^{\mathrm{cam}}
\end{equation}
where $\mathbf{z}_t^{\mathrm{imu}}$ and $\mathbf{z}_t^{\mathrm{cam}}$ are estimated quaternions by the IMU and camera, $\mathbf{v}_t^{\mathrm{imu}} \sim \mathcal{N}\left(0, R_{\mathrm{imu}}\right)$ and $\mathbf{v}_t^{\mathrm{cam}} \sim \mathcal{N}\left(0, R_{\mathrm{cam}}\right)$ are small measurement noises.

\textbf{Prediction Stage} The prediction is running with the same frequency as the IMU data reading. Let the $\hat{\mathbf{x}}_{t \mid t}$ be the current state estimated in the current time step, $\hat{\mathbf{x}}_{t+k \mid t}$ be the state after $k$ steps estimated in the current time step, $P_{t \mid t}$ be the current state covariance estimated in the current time, $P_{t+k \mid t}$ be the covariance after $k$ steps estimated in the current time,
\begin{equation}
    \begin{gathered}
\hat{\mathbf{q}}_{t+1 \mid t}=\hat{\mathbf{q}}_{t \mid t} \otimes \delta \mathbf{q}_t\\
\hat{\mathbf{b}}_{t+1 \mid t}=\hat{\mathbf{b}}_{t \mid t}
\end{gathered}
\end{equation}
Assuming small rotation increments, we have
\begin{equation}
    \hat{\mathbf{x}}_{t+1\mid t}=f\left(\hat{\mathbf{x}}_{t \mid t}\right)=\left[\begin{array}{c}
\hat{\mathbf{q}}_{t \mid t} \\
\hat{\mathbf{b}}_{t \mid t}
\end{array}\right]=\hat{\mathbf{x}}_{t\mid t}
\end{equation}

The state covariance is updated by
\begin{equation}
    P_{t+1 \mid t}=F_t P_{t \mid t} F_t^{\top}+Q_t
\end{equation}
where $F_t$ is the Jacobian of $f(\mathbf{x})$ with respect to $\mathbf{x}$, $F_t \approx I$, $Q_t$ is the process noise covariance.

\textbf{Update Stage} Update stage is followed with the prediction stage. In general, when the measurement received, update the related states and parameters by
\begin{equation}
    \begin{gathered}
K_t=P_{t \mid t-1} H_t^{\top}\left(H_t P_{t \mid t-1} H_t^{\top}+R\right)^{-1} \\
\hat{\mathbf{x}}_{t \mid t}=\hat{\mathbf{x}}_{t \mid t-1}+K_t\left(\mathbf{z}_t-h\left(\hat{\mathbf{x}}_{t \mid t-1}\right)\right) \\
P_{t \mid t}=\left(I-K_t H_t\right) P_{t \mid t-1}
\end{gathered}
\label{eq:EKF_update}
\end{equation}

The measurement function for the IMU is
\begin{equation}
h^{\mathrm{imu}}\left(\mathbf{x}_t\right)=\mathbf{q}_t \otimes\left[1, \frac{1}{2} \mathbf{b}_t\right]
\end{equation}

To perform the EKF update for IMU, linearizing the measurement function $h^{\mathrm{imu}}$ get
\begin{equation}
    H_t^{\mathrm{imu}}=\left.\frac{\partial h^{\mathrm{imu}}}{\partial \mathbf{x}}\right|_{\hat{\mathbf{x}}_{t \mid t-1}}
\end{equation}
which is then substitute with the $R$ in Eq. \ref{eq:EKF_update} to update states.

To perform the EKF update for camera, as $h^{\mathrm{cam}}\left(\mathbf{x}_t\right)=\mathbf{q}_t$, its Jacobian will be
\begin{equation}
    H_t^{\mathrm{cam}} \approx\left[\begin{array}{ll}
I_{4\times4} & 0_{4 \times 3}
\end{array}\right]
\end{equation}

To perform the EKF update for both camera and IMU, the measurement model can be written as
\begin{equation}
    \mathbf{z}_t=\left[\begin{array}{c}
\mathbf{\mathrm{z}}_t^{\mathrm{cam}} \\
\mathbf{\mathrm{z}_t^{\mathrm{imu}}}
\end{array}\right]=\left[\begin{array}{ll}
H^{\mathrm{cam}}_t \\
H^{\mathrm{imu}}_t
\end{array}\right]\mathbf{\mathrm{x}}_t+\left[\begin{array}{ll}
\mathbf{\mathrm{v}}^{\mathrm{cam}}_t \\
\mathbf{\mathrm{v}}^{\mathrm{imu}}_t
\end{array}\right]
\end{equation}
while $R=diag(R^{\mathrm{cam}},R^{\mathrm{imu}})$.

Finally, to maintain the unit norm of the quaternions, 
\begin{equation}
    \hat{\mathbf{x}}_{t \mid t} \leftarrow \frac{\hat{\mathbf{x}}_{t \mid t}}{\left\|\hat{\mathbf{x}}_{t \mid t}\right\|}
\end{equation}

\subsection{Adaptation to Latency}

\begin{figure}[!t] 
    \centering
    \includegraphics[width=\linewidth]{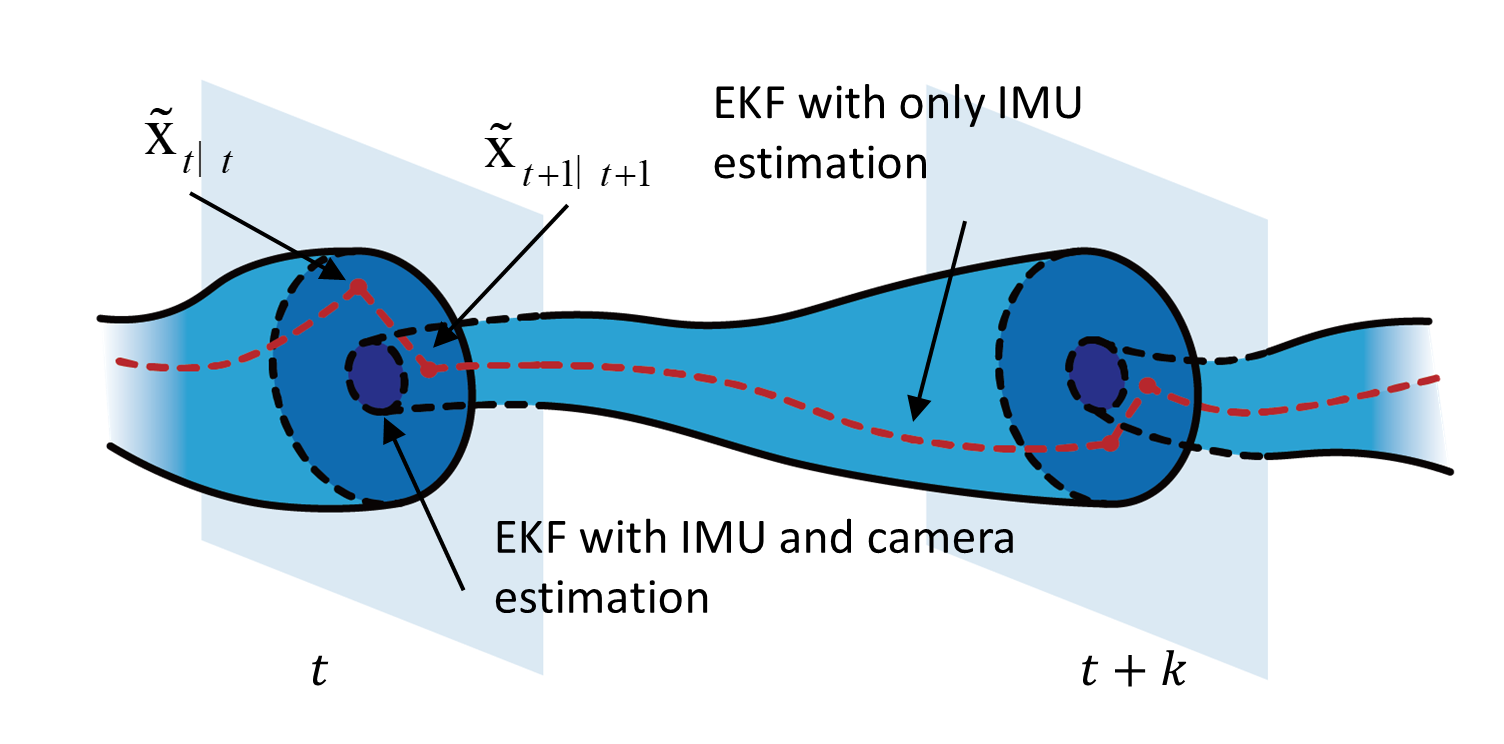}
    \caption{
        Illustration of visual-inertial updates within the EKF. Due to drift bias in IMU measurements, the inertial estimate diverges over time. Although visual estimation has latency, retrospective corrections can be applied to states from $k$ steps before the current time, allowing recomputation of the observation trajectory with higher accuracy.
    }
    \label{fig:funnel}
\end{figure}

\begin{figure*}[!t] 
    \centering
    \includegraphics[width=\linewidth]{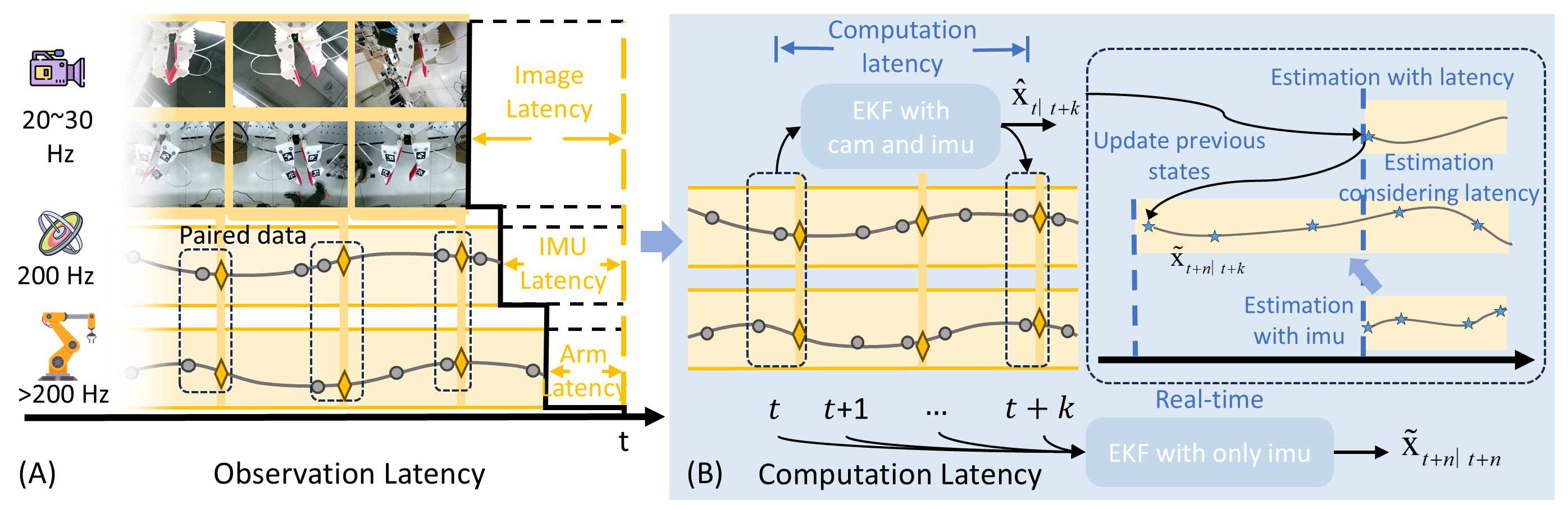}
    \caption{
        The latency free EKF. There are two sourses of latency, which are observation latency and computation latency. (A) shows observation latency between different input for attaining proprioception. (B) Computation latency is the major latency during EKF estimation. Therefore, we use IMU to conduct update and get real-time output, while use visual-inertial fusion to update states from $\hat{\mathbf{x}}_{t\mid t+k}$ to the current state $\tilde{\mathbf{x}}_{t+k \mid t+k}$. The computational latency is around 1s.
    }
    \label{fig:latency}
\end{figure*}

The proprioception, as a low-level perception feedback signal, requires a real-time update frequency. However, the visual-based algorithm always lack real-time capability, especially when running under learning-based algorithms. Therefore, to solve this problem, while maintaining the advantage of EKF's improved estimation robustness and accuracy, we revised the EKF to a latency free method.

We use real-time IMU data to continuously predict and update the pose trajectory. When a delayed visual frame arrives at timestep $t$ and this frame's inertial-visual EKF estimation arrives at timestep $t+k$, the EKF retrospectively refines the entire estimation from $t$ to $t+k$, where $k$ is the latency. This approach yields a more accurate estimation at $t+k$ than relying solely on inertial data (Fig. \ref{fig:funnel}) as below.

\begin{equation}
    \begin{gathered}
            \hat{\mathbf{x}}_{t+1\mid t+k}^{-}=\tilde{\mathbf{x}}_{t\mid t+k}\\
            \hat{P}_{t+1 \mid t+k}^{-}=F_t \tilde{P}_{t \mid t+k} F_t^{\top}+Q_t \\
            \hat{K}_{t+1\mid t+k}=\hat{P}_{t+1 \mid t+k}^{-} \tilde{H}_{t+1}^{\top}\left(\tilde{H}_{t+1} \hat{P}_{t+1 \mid t+k}^{-} \tilde{H}_{t+1}^{\top}+R\right)^{-1} \\
\hat{\mathbf{x}}_{t+1 \mid t+k}=\hat{\mathbf{x}}_{t+1\mid t+k}^{-}+\hat{K}_{t+1}\left(\overline{\mathbf{z}}_{t+1}^{\mathrm{imu}}-\tilde{H}_{t+1}^{\mathrm{imu}}\hat{\mathbf{x}}_{t+1\mid t+k}^{-}\right) \\
\tilde{P}_{t+1 \mid t+k}=\left(I-\hat{K}_{t+1\mid t+k} \tilde{H}_{t+1}^{\mathrm{imu}}\right) \hat{P}_{t+1 \mid t+k}^{-}
    \end{gathered}
    \notag
\end{equation}
where $\hat{\mathbf{x}}_{t \mid t+k}$, $\hat{P}_{t \mid t+k}$, and $\hat{K}_{t \mid t+k}$ are the visual-inertial state, covariance, and Kalman gain at timestep $t$ estimated at $t+k$. Based on this, we can refit all the prediction and updates using inertial information by 
\begin{equation}
    \begin{gathered}
            \tilde{\mathbf{x}}_{t+1\mid t}=\hat{\mathbf{x}}_{ t\mid t}\\
            \tilde{P}_{t+1 \mid t}=F_t \hat{P}_{t \mid t} F_t^{\top}+Q_t \\
            \tilde{K}_{t+1}=\tilde{P}_{t+1 \mid t} \tilde{H}_t^{\top}\left(\tilde{H}_t \tilde{P}_{t+1 \mid t} \tilde{H}_t^{\top}+R\right)^{-1} \\
\tilde{\mathbf{x}}_{t+1 \mid t+1}=\tilde{\mathbf{x}}_{t+1 \mid t}+\tilde{K}_{t+1}\left(\overline{\mathbf{z}}_{t+1}^{\mathrm{imu}}-\tilde{H}_{t+1}^{\mathrm{imu}}\tilde{\mathbf{x}}_{t+1 \mid t}\right) \\
\tilde{P}_{t+1 \mid t+1}=\left(I-\tilde{K}_{t+1} \tilde{H}_{t+1}^{\mathrm{imu}}\right) \tilde{P}_{t+1 \mid t}
    \end{gathered}
\end{equation}

\subsection{Solve the Configuration Space}

Given the pose as rotation matrix, we need to solve the joint angles of the dexterous hand \cite{slabaugh1999computing}, which can be formulated as solving
\begin{equation}
\overline{R}(\mathbf{z}_t) = R_i(\theta) R_j R_k(\phi) 
\end{equation}
where $i,j,k \in \{\mathrm{x, y, z}\}$, $R_j$ is a constant rotation matrix, $\theta$ and $\phi$ are the joint angles to be solved. To avoid dual solutions, we have the algorithm \ref{algo:rot2eul}.

\begin{algorithm}[!t]
\caption{Solve the Unique Configuration Space from Rotation Matrix}
\begin{algorithmic}[1]
\REQUIRE $\mathbf{R}_t, \alpha_{t-1},\beta_{t-1},\gamma_{t-1}\quad\triangleright$rotation matrix, Euler angles
\ENSURE Euler angles $\alpha_t, \beta_t, \gamma_t$ is uniquely attained given $\mathbf{R}_t=Z_\alpha Y_\beta X_\gamma$ 
\IF{$R_{31}\neq \pm 1$}
    \STATE $\beta_{t,1} = -\arcsin{R_{31}}$
    \STATE $\beta_{t,2} = \pi + \arcsin{R_{31}}$
    \IF{$\begin{vmatrix}
        \beta_{t-1}-\beta_{t,1}
    \end{vmatrix}\leq\begin{vmatrix}
        \beta_{t-1}-\beta_{t,2}
    \end{vmatrix}$}
        \STATE $\beta_t \leftarrow \beta_{t,1}$
    \ELSE
        \STATE $\beta_t \leftarrow \beta_{t,2}$
    \ENDIF
    \STATE $\alpha_t\leftarrow\text{atan2}(\frac{R_{21}}{\cos{\beta_t}},\frac{R_{11}}{\cos{\beta_t}})$
    \STATE $\gamma_t \leftarrow \text{atan2}(\frac{R_{32}}{\cos{\beta_t}},\frac{R_{33}}{\cos{\beta_t}})$
\ELSE
    \STATE $\alpha_t \leftarrow \alpha_{t-1}$
    \IF{$\beta_t=\frac{\pi}{2}$}
        \STATE $\gamma_t\leftarrow\alpha_t+\text{atan2}(R_{12},R_{13})$
    \ELSE
        \STATE $\gamma_t\leftarrow-\alpha_t+\text{atan2}(-R_{12},-R_{13})$
    \ENDIF
\ENDIF
\end{algorithmic}
\label{algo:rot2eul}
\end{algorithm}

\section{Experimental Verification} 
\label{sec:experiments}

In this section, we validate the feasibility and reliability of using IMUs to measure joint angles. We compare the performance of three pose estimation approaches: \textit{inertial-only}, \textit{visual-only}, and \textit{visual-inertial fusion}. Results demonstrate that our method exhibits accurate dynamic estimation and robust performance in motion-rich scenarios.

\begin{table}[htbp]
\centering
\caption{First-order Drift (in $\times 10^{-4} { }^\circ$ /s) and angular variance}
\label{tab:drift_coeff}
\begin{tabular*}{\columnwidth}{@{\extracolsep{\fill}}lcccccc}
\hline
\textbf{Pose} & \textbf{\#1} & \textbf{\#2} & \textbf{\#3} & \textbf{\#4} & \textbf{\#5} & \textbf{\#6} \\
\hline
\textbf{Roll}  & $-1$ & $33$  & $0$ & $-24$ & $0$ & $0$ \\
\textbf{Pitch} & $0$  & $0$ & $0$ & $0$ & $0$ & $1$ \\
\textbf{Yaw}   & $-8$ & $-36$ & $-1$ & $0$  & $-6$ & $1$ \\
\hline \hline
\textbf{Var (deg$^2$)} & 0.21 & 8.21 & 0.38 & 6.98 & 0.32 & 2.44 \\
\hline
\end{tabular*}
\end{table}

\subsection{IMU System Validation}

\begin{figure}[!t] 
    \centering
    \includegraphics[width=\linewidth]{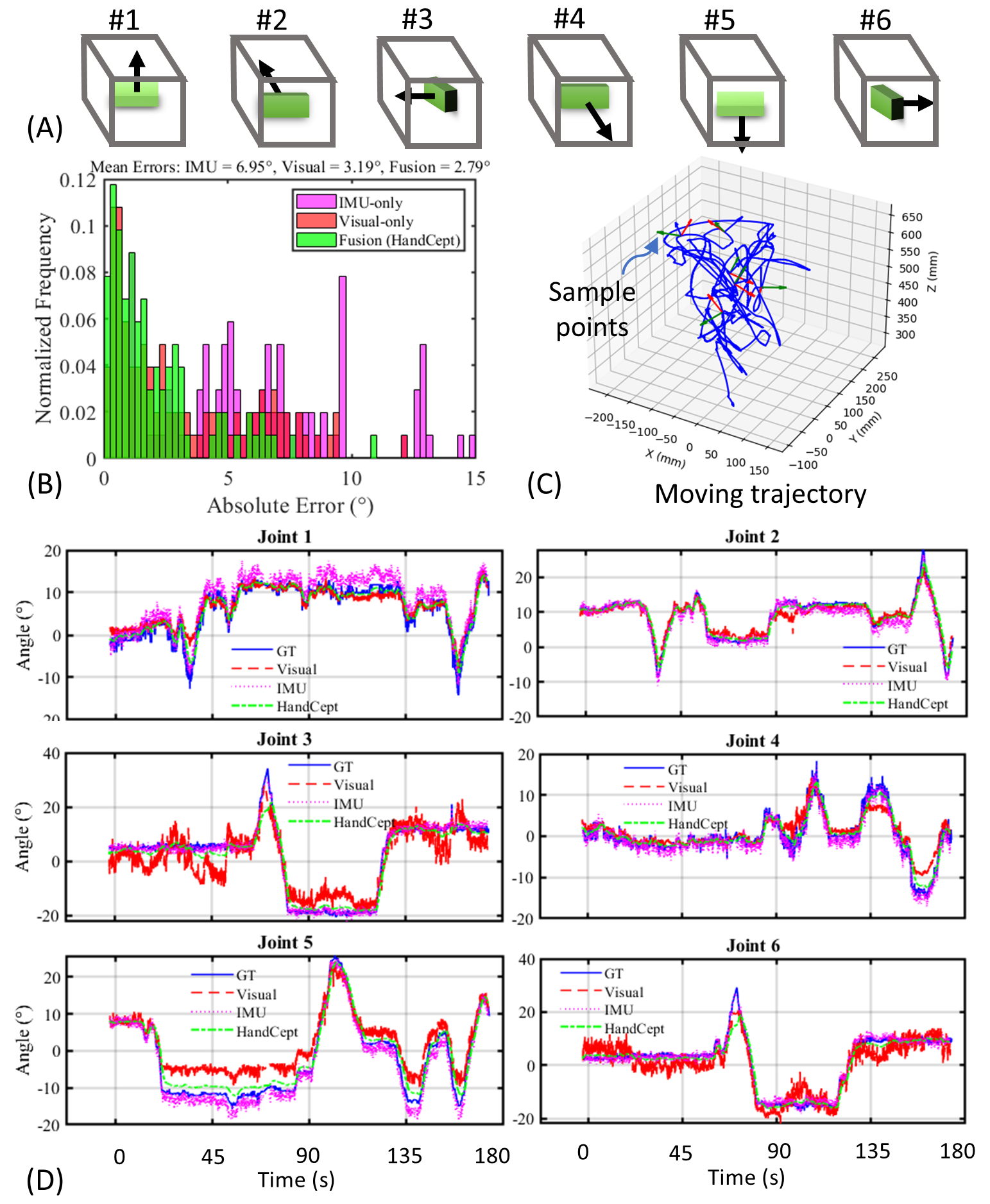}
    \caption{
Experimental results for free-space motion. (A) Placement and z-axis orientations used during IMU system verification. (B) Error distribution in multiple tests. (C) Trajectory of the robot arm's end-effector under random teleoperation for both position and rotation. The coordinate frames represent sampled end-effector poses. (D) Comparison of joint angle estimation. Joints 1, 2, and 3 correspond to the lateral, proximal, and distal joints of the left finger, respectively, while joints 4, 5, and 6 correspond to those right fingers.
    }
    \label{fig:exp}
\end{figure}

The static stability of the IMU measuring and the uniformity of 9-axis IMUs are validated in this section. The uniformity measures in which extent can we use a common base frame for different 9-axis IMUs. A common base frame will greatly reduce the difficulty in calibration for multiple IMUs systems.

\subsubsection{Stability of Inertial measurement}

In the static stability experiment, four IMUs were placed in six different orientations (Fig.\ref{fig:exp}A). After calibration under high-confidence condition, data were collected continuously over a two-hour period. The mean first-order drift (in degrees per second) across different chips is summarized in Table \ref{tab:drift_coeff}. Results show minimal drift following application of a normal Kalman filter, and no significant directional bias is observed.

\subsubsection{Uniformity of 9-axis IMU}

Similar to the stability experiment, the uniformity experiment placed 4 IMUs in those six directions \ref{fig:exp}A. Considering quaternion is continuous in its range, thus reasonable to get interpolation, we use quaternion to get the mean variance and avoid the gimbal lock. Let each unit quaternion measurement be represented as \( \mathbf{q}_i \in \mathbb{H} \), for \( i = 1, 2, \dots, N \), where \( N \) is the number of measurements for a given pose. The mean quaternion \( \mathbf{q}_{\text{mean}} \) is attained by minimizing the sum of squared geodesic distances on the rotation group SO(3),
\[
\mathbf{q}_{\text{mean}} = \arg\min_{\mathbf{q}} \sum_{i=1}^N d_{\mathrm{SO}(3)}^2(\mathbf{q}, \mathbf{q}_i)
\]

Then, the angular variance are obtained
\begin{equation}
    \begin{gathered}
        \delta \mathbf{q}_i = \mathbf{q}_i \cdot \mathbf{q}_{\text{mean}}^{-1} \\
        \theta_i = 2 \cdot \arccos(|w_i|) \\
        \text{Var}(\theta) = \frac{1}{N} \sum_{i=1}^N \left( \theta_i - \bar{\theta} \right)^2
    \end{gathered}
\end{equation}
where $w_i$ is the scalar part of $\delta \mathbf{q}_i$, $\text{Var}(\theta)$ is the variance. The BNO085 exhibits a typical heading drift of approximately $0.16 ^{\circ} $ per minute when using magnetometer-assisted sensor fusion. An experimental mean variance in different direction are shown in Table. \ref{tab:drift_coeff}. It's obvious that the variance is very small therefore is reasonable to use a common base for different 9-axis IMUs. The greatest variance happens when the z-axis is perpendicular to the gravity direction.

\subsection{Configuration Estimation Results for Free-Space Motion}

We use our previously developed DexCo hand~\cite{zhou2024dexterous}, which lacks an onboard joint estimation system, to test HandCept. While earlier methods relied purely on inertial data and required complex calibration, they suffered from long-term drift during dynamic motion. Here, we validate the HandCept framework using three modes: IMU-only, visual-only, and visual-inertial joint angle estimation.

The experimental setup is illustrated in Fig.\ref{fig:overview}A. The robot arm and dexterous hand are teleoperated to freely explore the workspace. Fig.\ref{fig:overview}B and Fig.~\ref{fig:overview}C depict a typical deployment using one camera and the IMU system. For evaluation, an additional overhead camera is used to capture ground truth poses using ArUco markers.

During trials, the robot moves randomly throughout its workspace, ensuring diverse motion and viewpoint coverage. Three 20 mins trails are conducted and summarized to present the error distribution (Fig. \ref{fig:exp}B). Fig.\ref{fig:exp}C shows the end-effector's 3D trajectory, which spans approximately half of a spherical volume, with directional variation at fixed positions to test sensitivity. A comparative plot in Fig.\ref{fig:exp}D shows the ground truth trajectory alongside estimates from each method. HandCept demonstrates superior performance, effectively resisting drift and achieving higher accuracy than either single-modality approach.

\subsection{Configuration Estimation on Downstream Tasks}

\begin{figure}[!t] 
    \centering
    \includegraphics[width=\linewidth]{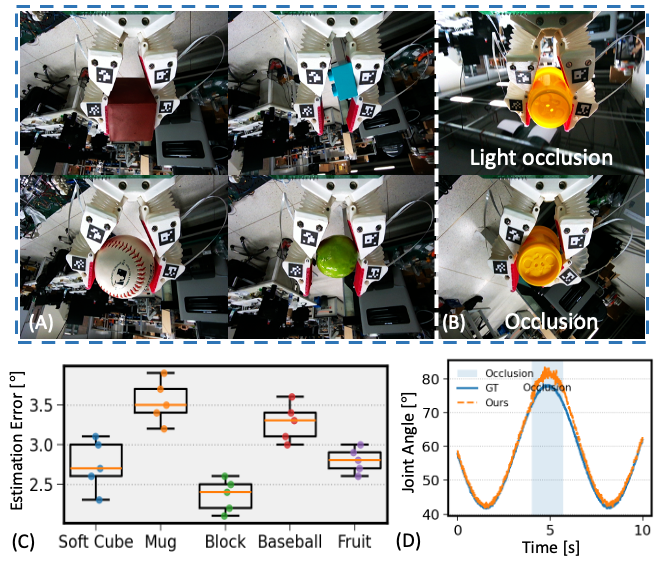}
    \caption{
Experimental results on grasping and in-hand manipulation. (A) Grasping on four different YCB objects. (B) In-hand manipulation on one YCB object. The right finger tip moves periodically in the experiment. (C) Box plots of joint-angle MAE (deg) over five YCB objects. Each object: five grasp trials; each trial value is the mean error across six joints. (D) Joint-angle trajectory over 10 s: ground truth vs. HandCept. Shaded 4.0–5.8 s marks a visual occlusion.
    }
    \label{fig:exp2}
\end{figure}

We evaluate HandCept on downstream tasks spanning two categories: \textbf{grasping} and \textbf{in-hand manipulation}. Across experiments we compare our method with ground truth (GT) data.

\textbf{Grasping.} We evaluated HandCept on a grasping benchmark using five YCB objects that vary in geometry and appearance \ref{fig:exp2}A. For each object, the DexCo hand repeatedly opened and closed until a stable grasp was achieved in five independent trials. In every trial we computed the mean absolute error (MAE) of the six joint angles and summarized the per–object distributions with a box plot (Fig.~\ref{fig:exp2}C). Across objects, HandCept produced tight error distributions with median MAE typically within $2^{\circ}$–$4^{\circ}$. Occasional outliers corresponded to transient self-occlusions or brief specular highlights, but fusion kept these errors bounded.

\textbf{In-hand manipulation.} We further tested HandCept during a 10 seconds in-hand manipulation sequence while recording both ground truth and HandCept estimates for a representative joint (Fig.~\ref{fig:exp2}D). A single visual occlusion was introduced between $t{=}4$,s and $t{=}5.8$ s as shown in Fig. \ref{fig:exp2}B. During occlusion, the estimate exhibits a small bias and slow drift, as expected from the failed visual updates; immediately after occlusion, the latency-compensated EKF re-synchronizes the estimate to the ground-truth trajectory with an exponential-like recovery, returning the error to its nominal band.

Together, the grasping and manipulation results show that HandCept delivers accurate and robust joint-angle estimation across object variation and dynamic motion, maintaining low median errors and quick recovery from visual dropouts—capabilities that are critical for closed-loop dexterous manipulation. However, occlusion affects the HandCept's capability. Because the visual part drifts greatly during occlusion, the estimation has a larger error during occlusion.


\section{Conclusion and Future Work}

This work introduces HandCept, a novel visual-inertial proprioception framework for dexterous hands—addressing accurate and latency-free dexterous hand proprioception, which is not previously resolved in the literature. HandCept estimates joint angles using a zero-shot visual-inertial approach, combining IMU readings and synthetic visual pose predictions within a latency-compensated EKF. We validate HandCept through two primary experiments. First, we demonstrate that 9-axis IMUs exhibit stable and uniform behavior across multiple orientations, enabling shared-base calibration and reducing setup complexity. Second, we show that HandCept achieves state-of-the-art accuracy, with joint angle errors maintained generally between $2^{\circ}$ to $4^{\circ}$ without observable drift.

In the future, we have four directions. Firstly, an improved real-time property both for the visual estimation algorithm and EKF estimation algorithm will be made. Secondly, visual estimation with a fine-tune technique should be introduced for a better stability and visual occlusion scenarios. Thirdly, based on the IMU, encoders, and camera, we will extend the current system with force sensing capability. Finally, adaptation to other dexterous robot platforms will be investigated.

\bibliographystyle{./bib/IEEEtran}
\bibliography{./bib/references}
\end{document}